\pdfoutput=1
\documentclass[11pt]{article}
\usepackage[preprint]{acl}
\usepackage{times}
\usepackage{latexsym}
\usepackage[T1]{fontenc}
\usepackage[utf8]{inputenc}
\usepackage{microtype}
\usepackage{inconsolata}
\usepackage{graphicx}
\usepackage{booktabs}
\usepackage{enumitem}
\usepackage{stfloats}

\title{Perspectives -- Interactive Document Clustering\\in the Discourse Analysis Tool Suite}

\author{
    ~Tim Fischer, ~Chris Biemann \\
    Language Technology Group, Department of Informatics, University of Hamburg, Germany \\
    \texttt{\{firstname.lastname\}@uni-hamburg.de}
}

\begin{document}

\maketitle
\begin{abstract}
This paper introduces \textit{Perspectives}, an interactive extension of the Discourse Analysis Tool Suite designed to empower Digital Humanities (DH) scholars to explore and organize large, unstructured document collections.
\textit{Perspectives} implements a flexible, aspect-focused document clustering pipeline with human-in-the-loop refinement capabilities.
We showcase how this process can be initially steered by defining analytical lenses through document rewriting prompts and instruction-based embeddings, and further aligned with user intent through tools for refining clusters and mechanisms for fine-tuning the embedding model.
The demonstration highlights a typical workflow, illustrating how DH researchers can leverage \textit{Perspectives}'s interactive document map to uncover topics, sentiments, or other relevant categories, thereby gaining insights and preparing their data for subsequent in-depth analysis.
\end{abstract}

\section{Introduction}
The Discourse Analysis Tool Suite\footnote{\href{https://github.com/uhh-lt/dats}{github.com/uhh-lt/dats}} (DATS) supports Digital Humanities (DH) researchers in data analysis by providing access to computational methods for large-scale, multimodal data.
Key functionalities include pre-processing, qualitative and quantitative analysis, and document retrieval (including keyword, metadata, and semantic search).
However, DATS users often struggle to explore and organize extensive corpora.
Keyword-based or semantic search functionalities are often insufficient and incompatible with a hermeneutic approach to conceptualization, and DATS lacks functionality for systematic structuring and a holistic overview.
Organizing documents, however, is vital for qualitative analysis and advanced DATS features, such as timeline analysis.

Document clustering offers a promising solution, particularly for discovering themes within a corpus, a key interest of DH scholars.
Modern clustering-based topic modeling \citep{grootendorst_bertopic_2022, angelov_topic_2024, reuter_gptopic_2024}, which leverages pre-trained language model embeddings, has gained considerable traction not only because it offers advantages in simplicity, modularity, and performance, but also because it presents interesting opportunities for steerability and interactivity.

Unsupervised clustering methods can present a \textit{take it or leave it} scenario for non-experts if the outputs do not align with their intent, rendering fully automatic methods ineffective.
For DH scholars, who often approach corpora with specific research questions in mind, the ability to interactively manipulate clustering is of great interest.
Such challenges are addressed by the Interactive Topic Modeling (ITM) community, which aims to achieve a controllable, collaborative human-machine process.

To address this gap, we introduce \textit{Perspectives}, an interactive document clustering extension for DATS.
Inspired by ITM and built on clustering-based neural topic modeling, we aim to make document clustering more accessible for DH scholars.
The system is centered around an interactive document map to support exploratory workflows and implements a clustering pipeline that can be steered and refined in multiple ways. 
While offering visualization and analysis tools, its primary goal within DATS is to help users organize their collections into tagged datasets, ready for use with other analytical tools. 
The contributions of this paper are:
\begin{enumerate}[leftmargin=0cm]
\itemindent1.2em
\itemsep0em  
    \item We introduce a flexible, aspect-focused clustering pipeline that combines (a) LLM-driven document rewriting to emphasize user-defined aspects, (b) instruction-steered embeddings to generate aspect-oriented document representations, and (c) few-shot fine-tuning of the embedding model to further align the representations with user intent.
    \item We evaluate the pipeline on multiple datasets, showing the effectiveness of aspect-oriented representations, document rewriting, and fine-tuning.
    \item We present \textit{Perspectives}, a human-in-the-loop document (HITL) clustering system that integrates the pipeline and provides an interactive 2D map for users to explore, validate, and directly manipulate clusters with established refinement operations.
\end{enumerate}

\section{Related Work}

The task of identifying topics in texts has evolved from traditional probabilistic models, such as Latent Dirichlet Allocation \citep{LDA}. Recent approaches, termed clustering-based neural topic modeling or topic discovery \citep{wu_survey_2024}, build on pre-trained language model embeddings. Top2Vec \citep{angelov2020top2vec} uses Doc2Vec \citep{doc2vec} embeddings, reduces their dimensionality with UMAP, clusters them using HDBSCAN, and identifies topics by finding words closest to cluster centroids. BERTopic \citep{grootendorst_bertopic_2022} follows a similar pipeline but employs a class-based TF-IDF mechanism to extract descriptive words for each identified cluster. Contextual Top2Vec \citep{angelov_topic_2024} refines this by representing documents with multiple segment vectors for fine-grained topic segmentation.

The topic modeling community has long recognized the need to make topic modeling more user-centric.
Seed wordlists \citep{jagarlamudi-etal-2012-seedwords}, seed documents \citep{grootendorst_bertopic_2022}, or seed sentences \citep{fang_user-centered_2023} can guide initial topic formation.
Further, interactive topic modeling systems address alignment with user intent by providing operations such as merging, splitting, or deleting topics, as well as adding or removing documents/words from specific topics \citep{fang_user-centered_2023}.
\citet{seelman_labeled_2024} proposes to update topic models by embedding user-assigned word labels to topics.
GPTopic \citep{reuter_gptopic_2024} leverages LLMs to build an accessible chat interface to explore and refine topics through natural language.

The field of interactive (semi-supervised) clustering also acknowledges the value of integrating expert knowledge.
While prior methods, such as hand-picked seed points \citep{basu2002seeding} or pairwise constraints \citep{basu2002seeding}, offered control, they required extensive user feedback.
\citet{viswanathan_large_2024} demonstrated that LLMs can guide semi-supervised clustering with minimal feedback by generating and encoding task-specific keyphrases for document representation.

Several systems facilitate the interactive exploration of document collections or topics.
Nomic Atlas\footnote{\href{https://atlas.nomic.ai}{atlas.nomic.ai}} is a notable commercial platform that offers high-performance 2D scatter-plot visualizations of document embeddings, enabling users to explore automatically extracted hierarchical topic clusters and interact via filtering and a chat assistant.
Related academic systems innovate with HITL focus, such as direct integration into the labeling workflow \citep{seelman_labeled_2024}, tracking model changes \citep{fang_user-centered_2023}, or RAG-based question-answering \citep{reuter_gptopic_2024}.

Effective visualization helps explore and understand clusters.
Standard techniques, such as top words, word clouds, document-cluster distributions, and 2D embedding projections \citep{wu_survey_2024}, are all available in our extension.

While inspired by existing platforms, such as Nomic Atlas, \textit{Perspectives} distinguishes itself by offering a user-centric HITL workflow tailored for DH: Users define the analytical lens before clustering and iteratively refine the outcomes afterward, with visualizations serving as a central component of this interactive sense-making process. 

\section{Envisioned Workflow}

\begin{figure*}[t]
\centering
    \frame{\includegraphics[width=1\textwidth]{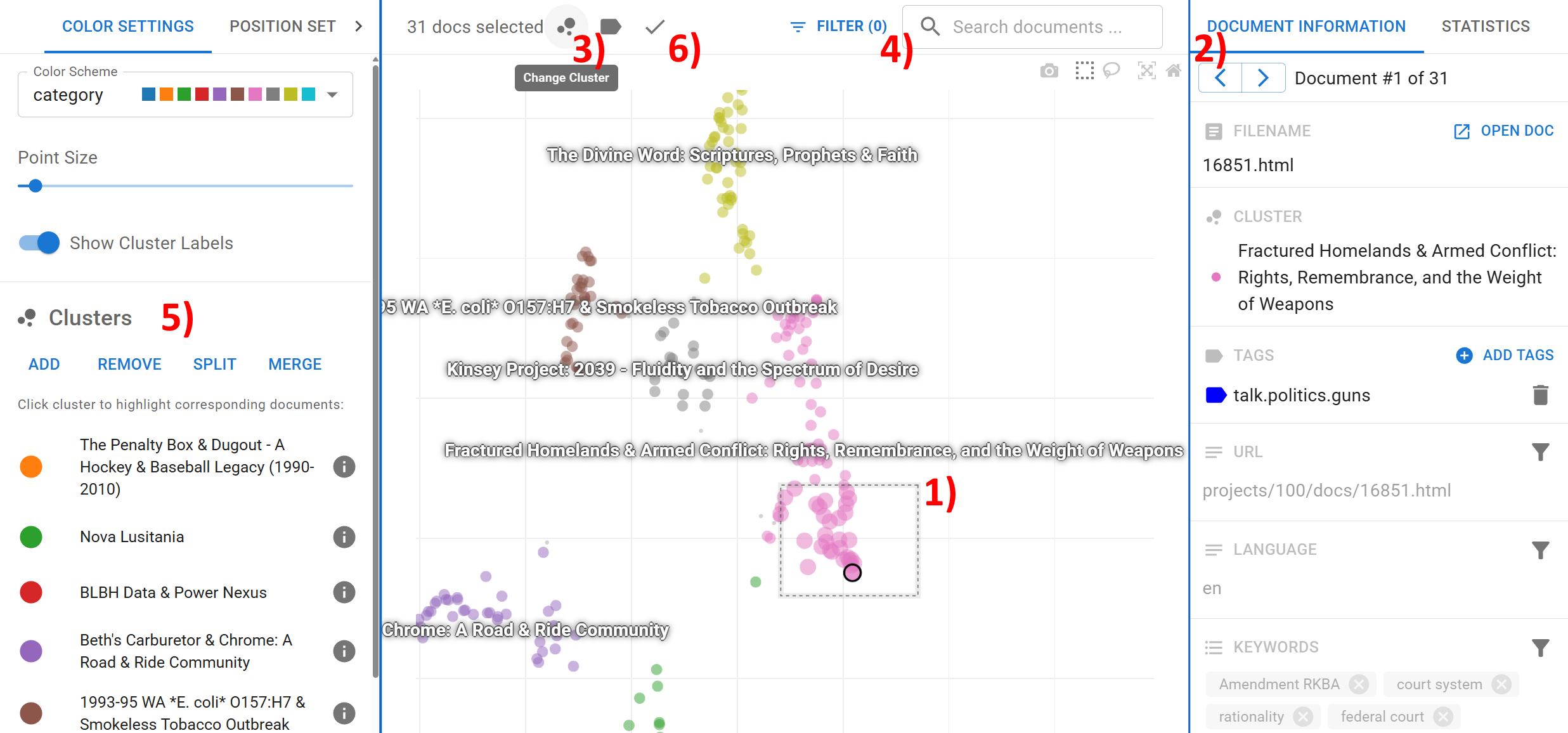}}
    \caption{\textit{Perspectives'} document map. Center: interactive scatter plot of documents colored by cluster. Hovering over a document previews its content. Left: settings \& refinement operations, including the model fine-tuning operation. Right: statistics \& information about selected documents. Top: toolbar with search \& filtering. The UI is inspired by popular clustering interfaces such as Nomic Atlas, leveraging a familiar design to accelerate adoption.}
\label{fig:ui}
\end{figure*}

This section illustrates an idealized workflow using \textit{Perspectives}, following Alice, a DH researcher investigating public discourse on climate change in Germany. 
Her goal is to understand major topics and the general sentiment towards climate change.

Alice uploads her corpus to DATS and opens the new \textit{Perspectives} extension.
To discover topics, she provides a \textit{topic discovery} instruction to guide document embeddings (e.g., "Identify the topic").
The system generates an interactive 2D document map and a dashboard that provides a high-level summary, including clusters with LLM-generated names, keywords, and example documents, their relative frequencies, and a similarity matrix showing inter-cluster relationships. 
Alice navigates the map (see Figure \ref{fig:ui}), panning, zooming, and inspecting the visually distinct clusters.
She identifies broad topics but notices that a theme like "Climate Protests" is missing.

Alice uses selection tools (1) to investigate documents of specific clusters.
Linked statistic and information panels (2) provide aggregated details, such as frequent keywords and named entities.
She spots misclassified articles, correcting them with the \textbf{Change Cluster} (3) function.
To create her expected "Climate Protests" topic, she selects relevant documents (found via integrated search or filtering, 4) and uses the \textbf{Add cluster} function.
If a topic is too broad, she uses the \textbf{Split Cluster} function to compute more granular sub-clusters.
Conversely, she can \textbf{Merge Clusters} if several clusters represent a single cohesive theme (5).

With all topics identified, Alice improves their definitions by reviewing representative documents and marking them with the \textbf{Accept Cluster Assignment} function (6).
Having provided minimal feedback, she triggers the \textbf{Refine Model} process, which fine-tunes the embedding model.
The map reloads, showing more distinct and coherent clusters.
Satisfied with her thematic map, Alice explores sentiment.
She creates a new perspective, this time providing a \textit{sentiment analysis} instruction and an optional document rewriting prompt (e.g., "Summarize sentiment towards climate change").
This generates a new map visualizing sentiment-based clusters, which she can explore and refine using the same interactive tools.

Finally, Alice exports her refined topics and sentiments from \textit{Perspectives} as tags back into her main project, enabling further analysis tools within the DATS ecosystem.

\section{Interactive Clustering}
\label{sec:interactive-clustering}
\begin{figure}[t]
\centering
    \includegraphics[width=.4\textwidth]{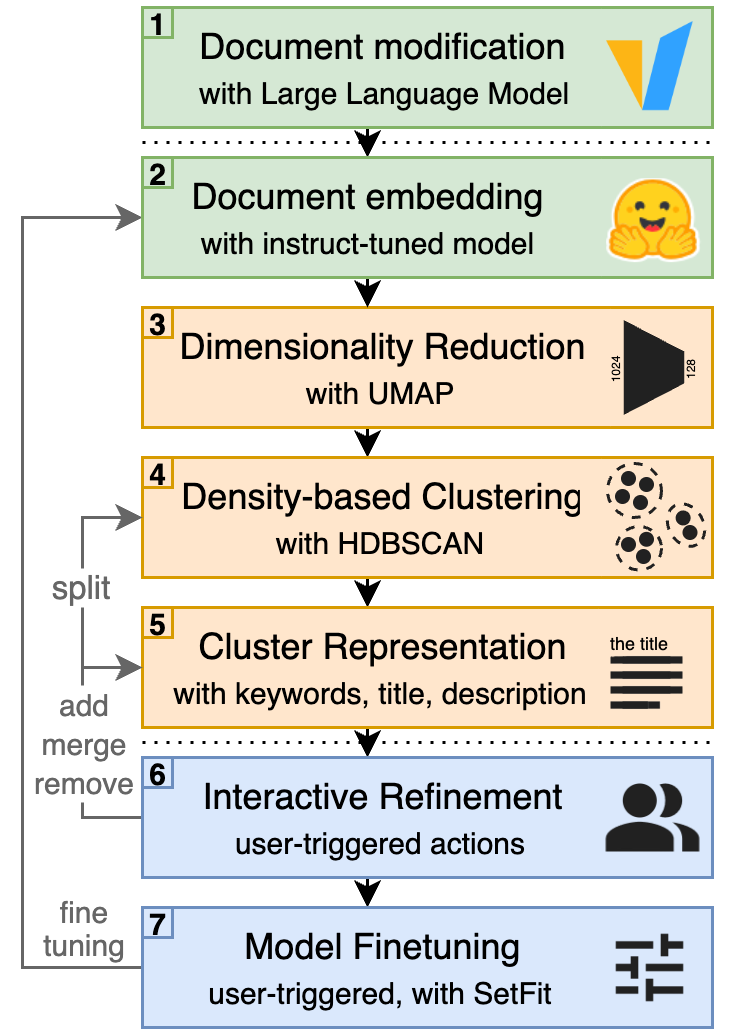}
    \caption{The proposed interactive clustering pipeline. The initial clustering process is guided by providing rewriting and embedding instructions (green) to focus the document representations on user-defined aspects. The established core pipeline (orange) identifies clusters and builds various textual representations. Users can post-process the clustering (blue) through refinement operations (grey), triggering some steps of the pipeline.}
\label{fig:pipeline}
\end{figure}

As illustrated by the workflow, \textit{Perspectives} facilitates interactive document clustering, empowering users to guide analysis towards specific aspects and iteratively refine categorizations.
This process begins with an aspect-focused clustering pipeline and continues through HITL refinement operations.
The pipeline is illustrated in Figure \ref{fig:pipeline}.

We propose two strategies to define the analytical lens and initially steer the clustering process (green).
First, we employ alternative document representations, in which an LLM rewrites documents based on a prompt, thereby modifying the content for clustering.
This is a one-time process per perspective.
Second, we use instruction-finetuned embedding models \citep{su-etal-2023-instructor} that generate aspect-oriented embeddings from an instruction, capturing specific semantic facets.

We continue with established steps (orange): reduce dimensionality with UMAP for improved clustering, then use HDBSCAN to group documents.
Cluster representations include (1) keywords from c-TF-IDF, (2) LLM-generated names and descriptions seeded by those keywords, (3) the cluster centroid (mean of document embeddings), and (4) representative documents identified via cosine similarity to the centroid.

Next, document embeddings (reduced to 2D using UMAP) are presented as colored dots on a map.
Here, users can iteratively post-process the clustering with several HITL refinement capabilities (blue; grey arrows).
Whenever clusters are manipulated (i.e., documents are added or removed), their representations (keywords, name, description, centroid, and representative documents) are recomputed.
The refinement operations are: 
\paragraph{Change Cluster} Documents are reassigned to an existing cluster or marked as an outlier. 
\vspace{-2mm}
\paragraph{Add Cluster (Documents)} Selected documents form a new cluster.
\vspace{-2mm}
\paragraph{Add Cluster (Text)} Provided name and description are embedded to define the cluster centroid. Documents are automatically reassigned if their cosine similarity to the new centroid exceeds their similarity to their current cluster's centroid.
\vspace{-2mm}
\paragraph{Merge Clusters} Two clusters are combined into a single new cluster.
\vspace{-2mm}
\paragraph{Remove Cluster} The cluster is deleted. Its documents' embeddings are compared to the centroids of all remaining clusters using cosine similarity and assigned to the most similar one.
\vspace{-2mm}
\paragraph{Split Cluster} HDBSCAN is applied only to the documents within the cluster, yielding a set of more granular sub-clusters, and the original cluster is removed.

Additionally, we propose fine-tuning the model to further align the document representations with user intent.
It is inspired by the findings of \citet{thielmann_human_2024}.
They demonstrated that with few (1-5) labeled documents per class, significant improvements in topic coherence can be achieved.
Perspectives operationalizes this by allowing users to \textit{accept cluster assignments}.
These labeled examples are used to create a contrastive dataset to fine-tune the embedding model with the SetFit\footnote{\href{https://huggingface.co/docs/setfit}{huggingface.co/docs/setfit}} library.
This adjusts the embedding space, pulling similar documents closer and pushing dissimilar ones apart (details are explained in Section \ref{sec:experiment-setup} and Appendix Section \ref{sec:appendix-experiment-details}).
After fine-tuning, the entire dataset is re-embedded and re-clustered, resulting in an updated map that reflects the improved cluster structure.

HDBSCAN inherently identifies outlier documents, visualized as distinctively colored (small, grey) points on the map.
They are considered during refinement operations.
For instance, when a new cluster is added, outliers can be assigned to it if their similarity is sufficient.

\section{Evaluation}
\label{sec:evaluation}
We evaluate the effectiveness of our additions to the established clustering pipeline, which aims to be steerable towards certain aspects.
We analyze initial guidance through (1) document rewrites and (2) instruction-tuned embeddings, as well as our proposed refinement operation of (3) model fine-tuning with few labeled data.

\subsection{KNN Accuracy Metric}
The goal of \textit{Perspectives} is to assist users in obtaining a solid document categorization using the 2D document map.
While topic modeling inspired our work, we refer to the extensive evaluation of \citet{angelov_topic_2024}, which demonstrated the superiority of clustering-based topic modeling approaches over traditional and neural methods.
Instead, we evaluate KNN accuracy, a standard metric for clustering interfaces. 
A good visualization should group documents with the same label and yield high classification accuracy within the visualization space \citep{pham_neural_2021}.
A KNN classifier is trained to classify texts based on their 2D coordinates, and accuracy is reported.
It mimics user behavior by deriving classes from dense clusters and can be seen as a proxy for end-to-end evaluation.

\subsection{Datasets}
To evaluate the effectiveness of steering and guidance, we selected various datasets that cover text materials (news articles, books, social media, and songs) and tasks (topic discovery, stance detection, frame analysis, and bias detection) similar to real-world research projects conducted by our project partners who actively use DATS. 

Amazon Product Reviews\footnote{\href{https://www.kaggle.com/datasets/mexwell/amazon-reviews-multi}{kaggle.com/datasets/mexwell/amazon-reviews-multi}} consists of 2,500 sampled reviews with 5 \textit{star} ratings and 15 \textit{product} categories.
Spotify Songtexts\footnote{\href{https://www.kaggle.com/datasets/devdope/900k-spotify}{kaggle.com/datasets/devdope/900k-spotify}} comprises 5,000 sampled song lyrics with 10 \textit{genres} and 5 \textit{emotions}.
20 Newsgroups includes 3,600 sampled posts on 20 \textit{topics}.
GVFC \citep{liu-etal-2019-detecting} contains 1300 news article summaries with 9 \textit{frames}.
Blurbs \citep{remus2019germeval} consists of 1,200 sampled German book \textit{blurbs} with 8 main genres.
Israel-Palestine \citep{ali2025socialmediapolarizationconflict} comprises 10,000 Reddit comments with 3 \textit{stances}.
News Bias\footnote{\href{https://www.kaggle.com/datasets/articoder/news-dataset-for-news-bias-analysis}{kaggle.com/datasets/articoder/news-bias-dataset}} contains 8,500 sampled news articles with 3 political \textit{biases}.
For detailed descriptions, see Appendix Section \ref{sec:appendix-dataset-details}.

\subsection{Models}
For evaluation, we exclusively use open-licensed models that run locally.
This approach ensures data privacy for our primary users (academic researchers) handling sensitive information.
It also ensures the system is deployable in resource-constrained environments, accommodating users with limited computing resources. We use Gemma 3 (27B-instruct) \cite{gemmateam2025gemma3technicalreport} for LLM-driven document rewriting and multilingual-e5-large-instruct (500M) \cite{wang2024multilinguale5textembeddings} for the instruction-tuned embedding model, selected due to its strong performance on the MTEB benchmark \citep{mteb}.
Further model details are provided in Appendix \ref{sec:appendix-model-details}.

\subsection{Experiment Setup}
\label{sec:experiment-setup}
First, we evaluate the impact of document rewriting and task-specific embeddings.
We use straightforward prompts to generate summaries and keyphrases of the original documents (see Appendix Table \ref{tab:embedding-instructions}).
Then, we use short instructions to compute task-specific document embeddings (see Appendix Table \ref{tab:modification-prompts}).
This results in 6 scenarios: original text (\textit{text}), summary (\textit{summ}), keyphrases (\textit{keyp}), embedded with (\textit{+inst}), and without instructions.
Second, we evaluate the impact of fine-tuning the embedding model with few labeled examples (\textit{2, 4, 8, 16-shot}), aiming to align the embedding space better with validated clusterings.
We use SetFit to tune the embedding model for one epoch on increasing amounts of examples sampled randomly from the training data.
The few labeled training examples are used to create a contrastive dataset with an equal number of positive and negative pairs.
Instead of updating all model parameters, we train LORA adapters \citep{hu2022lora} with a cosine similarity loss to speed up training and reduce hardware requirements.
The KNN accuracy metric is computed on document embeddings reduced to two dimensions with UMAP.
Few-shot experiments are repeated 10 times with different training sets.
The results are averaged across runs to mitigate fluctuations.
All experiments were conducted on a single A100 GPU (80GB).
All parameters, prompts, and training details are explained in Appendix \ref{sec:appendix-experiment-details}.
The code to reproduce the results is available\footnote{\href{https://github.com/uhh-lt/perspectives}{https://github.com/uhh-lt/perspectives}}.

\subsection{Results}

\begin{table*}[t]
\centering
\begin{tabular}{@{}llllllllll@{}}
\toprule
        & emotion        & genre          & product        & stars          & topic          & frames         & blurbs         & stance         & bias           \\ \midrule
text    & 45.10          & 27.20          & 36.72          & 47.54          & 71.17          & 59.78          & 73.84          & 51.19          & 47.20          \\
+inst   & \textbf{50.60} & 27.50          & 55.76          & \textbf{58.86} & \textbf{71.24} & 65.00          & 73.25          & 48.21          & 48.51          \\
keyphrases    & 49.15          & 34.85          & 54.64          & 46.34          & 65.70          & 60.34          & \textbf{76.73} & 61.63          & 50.76          \\
+inst   & 49.51          & \textbf{34.90} & 60.64          & 49.68          & 65.40          & 58.31          & 76.66          & 63.14          & 50.69          \\
summary    & 47.69          & 32.59          & 48.16          & 52.14          & 60.62          & 64.34          & 76.46          & 62.47          & 48.41          \\
+inst   & 48.04          & 32.64          & \textbf{61.18} & 54.02          & 61.91          & \textbf{65.91} & 74.80          & \textbf{68.74} & \textbf{52.53} \\ \midrule
2-shot  & 50.70          & 36.63          & 62.98          & 59.71          & 71.36          & 65.10          & 76.79          & 70.17          & 52.72          \\
4-shot  & 50.92          & 36.60          & 63.54          & 60.24          & 71.49          & 65.59          & 76.74          & 70.56          & 53.03          \\
8-shot  & 51.64          & 36.66          & 63.62          & 60.09          & 71.58          & 66.92          & 76.92          & 71.06          & 53.09          \\
16-shot & 52.07          & 37.02          & 64.04          & 61.27          & 72.15          & 67.27          & 77.85          & 71.51          & 54.09          \\ \bottomrule
\end{tabular}
\caption{Evaluation of document rewriting, instruction-based embeddings, and few-shot fine-tuning. Top: unsupervised (\textit{zero-shot}) results. Bottom: few-shot results using the best configuration (e.g, \textit{text+inst} for emotion \& topic).}
\label{table:evaluation}
\end{table*}

Table \ref{table:evaluation} presents our evaluation results (see Appendix Figure \ref{fig:detailed-evaluation} for a detailed breakdown.
Consistent with previous work, we observe that embeddings generated with instructions (\textit{+inst}) outperform those without in 8 out of 9 datasets we evaluated.
More interestingly, we observe that document rewrites (keyphrases or summaries) prove beneficial in 6 out of 9 datasets:
For example, \textit{summary+inst} improves product categorization accuracy by 24.46 points, stance detection by 17.55 points, and bias detection by 5.33 points.
For all other aspects, \textit{text+inst} yields the best performance.
Table \ref{table:evaluation} also highlights our best-performing few-shot results.
These scores were achieved using the best-performing \textit{zero-shot} configurations, for example \textit{summary+inst} for product, frames, stance, and bias, \textit{keyphrases+inst} for genre, and \textit{keyphrases} for blurbs.
While fine-tuning the embedding model with a few labeled examples consistently improved KNN accuracy, the gains were generally minor, in the range of 2-3 points, when compared to the \textit{zero-shot} setting.
\citet{thielmann_human_2024}, using 1-5 labeled examples per class, observed more significant improvements in topic coherence.
Still, our results suggest that incorporating more labeled examples could further enhance performance.

Our findings guided the implementation of \textit{Perspectives}:
Task-specific instructions are mandatory for creating a perspective, as they consistently yield superior performance.
The impact of document rewrites proved to be dataset-dependent.
Therefore, providing a document-rewriting prompt is optional but recommended.
To streamline the user experience, we integrated templates for common tasks, such as emotion, sentiment, and topic clustering.

\section{DATS Integration}
\textit{Perspectives} uses a React frontend with Plotly.js for efficient, interactive scatter-plot visualizations.
A FastAPI (Python) backend provides the functionalities through the asynchronous task manager RQ.
We implement GPU-powered RQ workers, which ensure UI responsiveness during intensive computations, such as embedding, clustering, and model fine-tuning.
We integrate our proposed clustering pipeline into DATS using the same parameters as in our experiments.

Fast response times are critical for interactive systems.
We achieve high performance for our LLM-driven functionalities (like name generation and document rewriting) by combining vLLM and LiteLLM\footnote{\href{https://www.litellm.ai/}{www.litellm.ai}, \href{https://docs.vllm.ai/en/latest/}{docs.vllm.ai}}.
This setup leverages batch processing, streaming, and prefix caching for rapid inference.
While the initial, optional document rewriting process may incur some latency depending on corpus size, it is an acceptable, one-time operation.
Subsequent refinement operations are significantly faster, as their processing time scales with the number of clusters.
Furthermore, the embedding model fine-tuning is also quick, requiring only one training epoch on a small set of labeled examples.

Using modern libraries such as Sentence Transformers and vLLM, switching models is straightforward.
This ensures that DATS stays up to date with recent advances in NLP and provides users with state-of-the-art models.

\section{Conclusion}
This paper introduced \textit{Perspectives}\footnote{
\href{https://dats.ltdemos.informatik.uni-hamburg.de/}{Demo: dats.ltdemos.informatik.uni-hamburg.de}}, an interactive extension to the open-source DATS platform designed to help DH scholars explore and organize large document collections.
We demonstrated an aspect-focused clustering pipeline that leverages LLM-based document rewriting and instruction-steered embeddings.
Central to \textit{Perspectives} is HITL refinement, providing a suite of operations to manipulate clusters and fine-tune the embedding model with minimal input.
This iterative approach, visualized on an interactive 2D map and supported by a dashboard, empowers users to create meaningful categorizations.
By integrating these capabilities, \textit{Perspectives} provides a powerful yet accessible tool that enables domain experts to leverage their knowledge effectively.
It transforms document clustering into a collaborative sense-making activity, prepares data for further analysis within DATS, and provides a valuable tool for user-centered DH research.
Future work will expand the UI's functionality, extend the scope of the extension to include multimodal data (images), and incorporate more labeled examples for few-shot scenarios.

\section*{Limitations}
During initial testing of \textit{Perspectives} with our project partners, a challenge in the model fine-tuning mechanism was identified:
The \textbf{Refine Model} operation requires re-embedding and re-clustering of the entire dataset, resulting in the loss of spatial understanding and disruption of the user's mental map as the 2D projection recalculates.
To mitigate this, we are currently developing a refinement history component.
This feature, presented as an interactive timeline at the bottom of the map, stores and allows navigation between all previous map states (e.g., pre- and post-refinement views).
We animate the transition between any two states, which helps users reconstruct their spatial understanding of the data.
This feature also addresses a request from project partners, allowing them to revert undesirable refinement operations.

Our experimental evaluation focused on a single LLM for document rewriting and a single embedding model.
Different choices of LLMs or embedding models could yield varying performance outcomes.
Furthermore, the few-shot training approach is sensitive to the selection of training samples.
To mitigate this high dependency, we conducted 10 runs with different sampled training sets.
However, an optimal combination of training examples may not have been discovered.
Our experiments also did not involve extensive prompt engineering for document rewriting, as we limited the prompts to summarization and keyphrase generation.
Exploring other prompts could potentially lead to further improvements. 

\bibliography{custom}

\begin{thebibliography}{23}
\providecommand{\natexlab}[1]{#1}

\bibitem[{Ali et~al.(2025)Ali, Abrar, Hossain, and Mridha}]{ali2025socialmediapolarizationconflict}
Hasin~Jawad Ali, Ajwad Abrar, S.~M.~Hozaifa Hossain, and M.~Firoz Mridha. 2025.
\newblock \href {https://arxiv.org/abs/2502.00414} {{Social media polarization during conflict: Insights from an ideological stance dataset on Israel-Palestine Reddit comments}}.

\bibitem[{Angelov(2020)}]{angelov2020top2vec}
Dimo Angelov. 2020.
\newblock \href {https://arxiv.org/abs/2008.09470} {{Top2Vec: Distributed Representations of Topics}}.

\bibitem[{Angelov and Inkpen(2024)}]{angelov_topic_2024}
Dimo Angelov and Diana Inkpen. 2024.
\newblock \href {https://doi.org/10.18653/v1/2024.findings-emnlp.790} {{Topic {Modeling}: {Contextual} {Token} {Embeddings} {Are} {All} {You} {Need}}}.
\newblock In \emph{{Findings of the {Association} for {Computational} {Linguistics}: {EMNLP} 2024}}, pages 13528--13539, Miami, FL, USA. {ACL}.

\bibitem[{Basu et~al.(2002)Basu, Banerjee, and Mooney}]{basu2002seeding}
Sugato Basu, Arindam Banerjee, and Raymond~J. Mooney. 2002.
\newblock {Semi-supervised Clustering by Seeding}.
\newblock In \emph{{Proceedings of the Nineteenth International Conference on Machine Learning}}, page 27–34, San Francisco, CA, USA. {Morgan Kaufmann Publishers}.

\bibitem[{Blei et~al.(2003)Blei, Ng, and Jordan}]{LDA}
David~M. Blei, Andrew~Y. Ng, and Michael~I. Jordan. 2003.
\newblock {Latent dirichlet allocation}.
\newblock \emph{The Journal of Machine Learning Research}, 3:993–1022.

\bibitem[{Conneau et~al.(2020)Conneau, Khandelwal, Goyal, Chaudhary, Wenzek, Guzm{\'a}n, Grave, Ott, Zettlemoyer, and Stoyanov}]{conneau-etal-2020-unsupervised}
Alexis Conneau, Kartikay Khandelwal, Naman Goyal, Vishrav Chaudhary, Guillaume Wenzek, Francisco Guzm{\'a}n, Edouard Grave, Myle Ott, Luke Zettlemoyer, and Veselin Stoyanov. 2020.
\newblock \href {https://doi.org/10.18653/v1/2020.acl-main.747} {{Unsupervised Cross-lingual Representation Learning at Scale}}.
\newblock In \emph{{Proceedings of the 58th Annual Meeting of the Association for Computational Linguistics}}, pages 8440--8451, Online. {ACL}.

\bibitem[{Fang et~al.(2023)Fang, Alqazlan, Liu, He, and Procter}]{fang_user-centered_2023}
Zheng Fang, Lama Alqazlan, Du~Liu, Yulan He, and Rob Procter. 2023.
\newblock \href {https://doi.org/10.18653/v1/2023.eacl-main.37} {{A {User}-{Centered}, {Interactive}, {Human}-in-the-{Loop} {Topic} {Modelling} {System}}}.
\newblock In \emph{{Proceedings of the 17th {Conference} of the {European} {Chapter} of the {Association} for {Computational} {Linguistics}}}, pages 505--522, Dubrovnik, Croatia. {ACL}.

\bibitem[{GemmaTeam(2025)}]{gemmateam2025gemma3technicalreport}
GemmaTeam. 2025.
\newblock \href {https://arxiv.org/abs/2503.19786} {{Gemma 3 Technical Report}}.

\bibitem[{Grootendorst(2022)}]{grootendorst_bertopic_2022}
Maarten Grootendorst. 2022.
\newblock \href {https://doi.org/10.48550/arXiv.2203.05794} {{{BERTopic}: {Neural} topic modeling with c-{TF}-{IDF}}}.

\bibitem[{Hu et~al.(2022)Hu, Shen, Wallis, Allen-Zhu, Li, Wang, Wang, and Chen}]{hu2022lora}
Edward~J Hu, Yelong Shen, Phillip Wallis, Zeyuan Allen-Zhu, Yuanzhi Li, Shean Wang, Lu~Wang, and Weizhu Chen. 2022.
\newblock \href {https://openreview.net/forum?id=nZeVKeeFYf9} {{Lo{RA}: Low-Rank Adaptation of Large Language Models}}.
\newblock In \emph{{International Conference on Learning Representations}}, Online.

\bibitem[{Jagarlamudi et~al.(2012)Jagarlamudi, Daum{\'e}~III, and Udupa}]{jagarlamudi-etal-2012-seedwords}
Jagadeesh Jagarlamudi, Hal Daum{\'e}~III, and Raghavendra Udupa. 2012.
\newblock \href {https://aclanthology.org/E12-1021/} {{Incorporating Lexical Priors into Topic Models}}.
\newblock In \emph{{Proceedings of the 13th Conference of the {E}uropean Chapter of the Association for Computational Linguistics}}, pages 204--213, Avignon, France. {ACL}.

\bibitem[{Le and Mikolov(2014)}]{doc2vec}
Quoc Le and Tomas Mikolov. 2014.
\newblock {Distributed representations of sentences and documents}.
\newblock In \emph{{Proceedings of the 31st International Conference on International Conference on Machine Learning}}, page 1188–1196, Beijing, China. {JMLR.org}.

\bibitem[{Liu et~al.(2019)Liu, Guo, Mays, Betke, and Wijaya}]{liu-etal-2019-detecting}
Siyi Liu, Lei Guo, Kate Mays, Margrit Betke, and Derry~Tanti Wijaya. 2019.
\newblock \href {https://doi.org/10.18653/v1/K19-1047} {{Detecting Frames in News Headlines and Its Application to Analyzing News Framing Trends Surrounding {U}.{S}. Gun Violence}}.
\newblock In \emph{{Proceedings of the 23rd Conference on Computational Natural Language Learning (CoNLL)}}, pages 504--514, Hong Kong, China. {ACL}.

\bibitem[{Muennighoff et~al.(2023)Muennighoff, Tazi, Magne, and Reimers}]{mteb}
Niklas Muennighoff, Nouamane Tazi, Loic Magne, and Nils Reimers. 2023.
\newblock \href {https://doi.org/10.18653/v1/2023.eacl-main.148} {{{MTEB}: Massive Text Embedding Benchmark}}.
\newblock In \emph{{Proceedings of the 17th Conference of the European Chapter of the Association for Computational Linguistics}}, pages 2014--2037, Dubrovnik, Croatia. {ACL}.

\bibitem[{Pham and Le(2021)}]{pham_neural_2021}
Dang Pham and Tuan M.~V. Le. 2021.
\newblock \href {https://doi.org/10.1007/978-3-030-86523-8_3} {{Neural {Topic} {Models} for {Hierarchical} {Topic} {Detection} and {Visualization}}}.
\newblock In \emph{{Machine {Learning} and {Knowledge} {Discovery} in {Databases}. {Research} {Track}}}, pages 35--51, Cham. {Springer International Publishing}.

\bibitem[{Remus et~al.(2019)Remus, Aly, and Biemann}]{remus2019germeval}
Steffen Remus, Rami Aly, and Chris Biemann. 2019.
\newblock {GermEval 2019 Task 1: Hierarchical Classification of Blurbs.}
\newblock In \emph{{Proceedings of the 15th Conference on Natural Language Processing (KONVENS 2019)}}, Erlangen, Germany. {German Society for Computational Linguistics \& Language Technology}.

\bibitem[{Reuter et~al.(2024)Reuter, Thielmann, Weisser, Fischer, and Säfken}]{reuter_gptopic_2024}
Arik Reuter, Anton Thielmann, Christoph Weisser, Sebastian Fischer, and Benjamin Säfken. 2024.
\newblock \href {https://doi.org/10.48550/arXiv.2403.03628} {{{GPTopic}: {Dynamic} and {Interactive} {Topic} {Representations}}}.

\bibitem[{Seelman et~al.(2024)Seelman, Zhang, and Boyd-Graber}]{seelman_labeled_2024}
Kyle Seelman, Mozhi Zhang, and Jordan Boyd-Graber. 2024.
\newblock \href {https://doi.org/10.48550/arXiv.2311.09438} {{Labeled {Interactive} {Topic} {Models}}}.

\bibitem[{Su et~al.(2023)Su, Shi, Kasai, Wang, Hu, Ostendorf, Yih, Smith, Zettlemoyer, and Yu}]{su-etal-2023-instructor}
Hongjin Su, Weijia Shi, Jungo Kasai, Yizhong Wang, Yushi Hu, Mari Ostendorf, Wen-tau Yih, Noah~A. Smith, Luke Zettlemoyer, and Tao Yu. 2023.
\newblock \href {https://doi.org/10.18653/v1/2023.findings-acl.71} {{One Embedder, Any Task: Instruction-Finetuned Text Embeddings}}.
\newblock In \emph{{Findings of the Association for Computational Linguistics}}, pages 1102--1121, Toronto, Canada. {ACL}.

\bibitem[{Thielmann et~al.(2024)Thielmann, Weisser, and Säfken}]{thielmann_human_2024}
Anton~F. Thielmann, Christoph Weisser, and Benjamin Säfken. 2024.
\newblock \href {https://aclanthology.org/2024.lrec-main.736/} {{Human in the {Loop}: {How} to {Effectively} {Create} {Coherent} {Topics} by {Manually} {Labeling} {Only} a {Few} {Documents} per {Class}}}.
\newblock In \emph{{Proceedings of the 2024 {Joint} {International} {Conference} on {Computational} {Linguistics}, {Language} {Resources} and {Evaluation}}}, pages 8395--8405, Torino, Italia. {ELRA and ICCL}.

\bibitem[{Viswanathan et~al.(2024)Viswanathan, Gashteovski, Gashteovski, Lawrence, Wu, and Neubig}]{viswanathan_large_2024}
Vijay Viswanathan, Kiril Gashteovski, Kiril Gashteovski, Carolin Lawrence, Tongshuang Wu, and Graham Neubig. 2024.
\newblock \href {https://doi.org/10.1162/tacl_a_00648} {{Large {Language} {Models} {Enable} {Few}-{Shot} {Clustering}}}.
\newblock \emph{Transactions of the Association for Computational Linguistics}, 12:321--333.
\newblock Place: Cambridge, MA Publisher: MIT Press.

\bibitem[{Wang et~al.(2024)Wang, Yang, Huang, Yang, Majumder, and Wei}]{wang2024multilinguale5textembeddings}
Liang Wang, Nan Yang, Xiaolong Huang, Linjun Yang, Rangan Majumder, and Furu Wei. 2024.
\newblock \href {https://arxiv.org/abs/2402.05672} {{Multilingual E5 Text Embeddings}}.

\bibitem[{Wu et~al.(2024)Wu, Nguyen, and Luu}]{wu_survey_2024}
Xiaobao Wu, Thong Nguyen, and Anh~Tuan Luu. 2024.
\newblock \href {https://doi.org/10.1007/s10462-023-10661-7} {{A survey on neural topic models: methods, applications, and challenges}}.
\newblock \emph{Artificial Intelligence Review}, 57(2):18.

\end{thebibliography}

\appendix

\section{Appendix}
\label{sec:appendix}

\subsection{Dataset Details}
\label{sec:appendix-dataset-details}
\paragraph{Amazon Product Reviews} Customer reviews organized by product category and star rating (proxy for sentiment). We use English split and consider reviews of the 15 most frequent categories. We sample 500 reviews per star for testing.
\vspace{-2mm}
\paragraph{Spotify Songtexts} Song lyrics categorized by \textit{emotion} and \textit{genre}. We consider the top 10 genres and top 5 emotions for English song texts. We sample 500 song texts per genre for testing.
\vspace{-2mm}
\paragraph{20 Newsgroups} A topic modeling dataset comprising 18000 newsgroups posts on 20 topics. We use 20\% of the articles for testing.
\vspace{-2mm}
\paragraph{GVFC} The Gun Violence Frame Corpus comprises news article headlines, annotated into 9 different frames.
We use an extension, which includes three-sentence summaries of 1300 articles. 
\vspace{-2mm}
\paragraph{German Blurbs} This dataset consists of 20k blurbs categorized into 8 main genres, each with its own fine-grained sub-genres. We sample 1200 blurbs for testing. We consider the main genre.
\vspace{-2mm}
\paragraph{Israel-Palestine} A stance detection dataset of 10k Reddit comments related to the Israel-Palestine conflict, categorized as Israel, Palestine, Neutral.
\vspace{-2mm}
\paragraph{News Bias} A dataset 8.5k news events retrieved from AllSides. It provides three views (left, center, right) for each news event. We sample 10k documents from the top 8 topics for testing.

\subsection{Experiment Parameters \& Libraries}
\label{sec:appendix-experiment-details}

\paragraph{Document Modification}
We use the instruction-tuned Gemma 3 (27B) model.
The vLLM and LiteLLM libraries handle this process for batch inference, with the \textit{max tokens to generate} parameter set to 2048; other sampling parameters, such as \textit{temperature} and \textit{top p}, remain at their default values.
Prompts used for document modification are listed in Table \ref{tab:modification-prompts}.

\paragraph{Document Embedding}
We use the Sentence Transformers library with \textit{batch size} set to 32 and \textit{normalize embeddings} set to true to generate aspect-oriented embeddings with the instruction-tuned \textit{multilingual-e5-large-instruct} (500M) embedding model.
Instructions used for document embedding are listed in Table \ref{tab:embedding-instructions}

\paragraph{Dimensionality Reduction}
A GPU-accelerated variant of UMAP from the cuML library reduces the dimensionality of resulting embeddings.
The \textit{distance metric} is cosine similarity, \textit{N neighbors} is 15, and \textit{min distance} is 0.0.
For evaluation, \textit{N components} is 2.
For production, it is 128.

\paragraph{KNN Accuracy Metric}
The k-nearest neighbors accuracy metric is computed using the sklearn library via 5-fold cross-validation, with the reported accuracies averaged across folds.

\paragraph{Density-based Clustering}
We use HDBSCAN for density-based clustering. The \textit{min samples} is set to 40, \textit{distance metric} is set to euclidean.

\paragraph{Cluster Representation}
c-TF-IDF extracts 50 keywords, Gemma 3 generates title \& description using constrained generation to ease parsing.

\paragraph{Interactive Refinement}
The split operation uses the same HDBSCAN parameters, other operations have no parameters.

\paragraph{Model Fine-tuning}
We use the SetFit library to fine-tune the embedding model in two stages.

For the first stage (one epoch), ten distinct few-shot training sets are created per dataset and shot count (2, 4, 8, 16) by randomly sampling \textit{num shots} examples per class.
They are constructed as contrastive datasets of document pairs, where a pair is deemed positive if both belong to the same class and negative otherwise.
Positive pairs are oversampled to balance the dataset, which enables substantial training even with limited labeled samples.
We use rank-stabilized LoRA adapters with a \textit{rank} of 24, an \textit{alpha} of 8, and a \textit{dropout} of 0.1, drastically reducing the number of trainable parameters.
The training objective is to minimize cosine similarity loss. 

The second stage (16 epochs) involves training a differentiable classification head on the updated embedding model.
We set the \textit{end-to-end training} parameter to true to further train the embedding model.
Since our focus is solely on the document embeddings, the classification head is discarded after training.

\begin{table*}[b!]
\begin{tabular}{@{}ll@{}}
\toprule
Dataset & Instruction                                                                                 \\ \midrule
emotion & Identify the main emotion expressed in the given (summary of a | keyphrases of a) song text \\
genre   & Identify the main genre of the given (summary of a | keyphrases of a) song text             \\
stars   & Identify the sentiment of the given (summary of an | keyphrases of an) Amazon review        \\
product & Identify the category of an Amazon product based on the review (summary | keyphrases) \\
topic   & Identify the topic or theme of the given (summary of a | keyphrases of a) news article      \\
frame   & Identify the framing of the given (summary of a | keyphrases of a) news article      \\
blurb   & Identifiziere das Genre, dass durch die (Zusammenfassung | Schlüsselbegriffe) des \\ 
        & Klappentextes beschrieben wird. \\
stance  & Identify the stance towards the Israel-Palestine conflict of the given (summary | keyphrases) \\ 
        & of a news article  \\
bias    & Identify the political leaning of the given (summary of a | keyphrases of a) news article      \\ \bottomrule
\end{tabular}
\caption{Instructions provided to the embedding model. We adopted them to resemble those of the original authors.}
\label{tab:embedding-instructions}
\end{table*}

\newpage

\subsection{Model Details}
\label{sec:appendix-model-details}

We utilized two primary models for the evaluation:

\paragraph{Gemma 3 (27B-instruct) for Document Rewriting}
This is a \textit{lightweight}, instruction-tuned LLM from Google.
It was trained on 14T tokens, features a large 128k context window, and supports over 140 languages.

\paragraph{multilingual-e5-large-instruct (500M) for Document Embedding}
This instruction-tuned embedding model was selected as one of the best-performing models on the clustering subtask of the MTEB benchmark \citep{mteb}.
It is initialized from XLM Roberta \citep{conneau-etal-2020-unsupervised}, continually trained on a mixture of multilingual datasets, and supports 100 languages.

\begin{figure*}[t]
\centering
    \includegraphics[width=\textwidth]{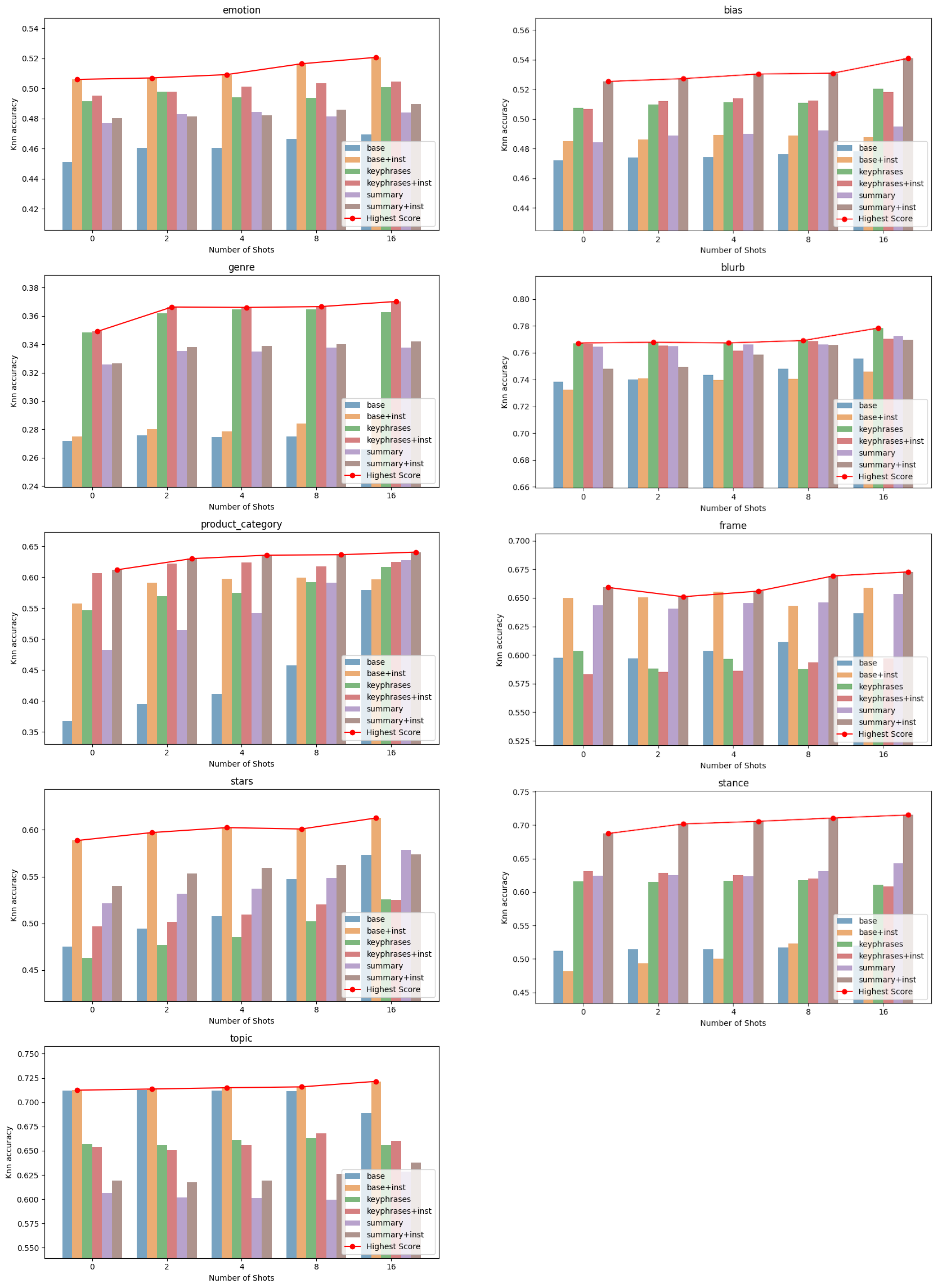}
    \caption{Detailed evaluation with increasing number of labeled examples per class.}
\label{fig:detailed-evaluation}
\end{figure*}

\begin{table*}[t]
\begin{tabular}{@{}ll@{}}
\toprule
Dataset             &   Prompt \\ \midrule
emotion-summary     &   Write a concise summary (maximum 5 sentences) that focuses on the emotional \\
                    &   tone of the following song lyrics. Analyze the lyrics to determine the main \\
                    &   emotion being conveyed and describe how it is expressed. \\
                    &   Conclude with an emotion categorization:                            \\
emotion-keyphrases  &   Generate keyphrases (max 5 phrases) that describe the emotional tone of \\
                    &   the following song lyrics. Focus on phrases that reflect the emotional tone, \\
                    &   mood, or feelings expressed in the lyrics. \\ 
genre-summary       &   Write a concise summary (maximum 5 sentences) that focuses on the genre of \\
                    &   the following song lyrics. Analyze the lyrics to determine the musical genre, \\ 
                    &   referencing stylistic elements, themes, or influences. \\
                    &   Conclude with a genre categorization: \\
genre-keyphrases    &   Generate keyphrases (max 5 phrases) that describe the musical genre of \\
                    &   the following song lyrics. Focus on phrases that reflect the genre's \\
                    &   characteristics, style, or influences.  \\
stars-summary       &   Write a concise summary (maximum 5 sentences) that focuses on the sentiment\\
                    &   of the following Amazon review. Analyze the review to determine the main  \\
                    &   sentiment being conveyed and describe how it is expressed. \\
                    &   Conclude with a sentiment categorization: \\
stars-keyphrases    &   Generate keyphrases (max 5 phrases) that describe the sentiment of the  \\
                    &   following Amazon review. Focus on phrases that reflect the sentiment, mood,  \\
                    &   or feelings expressed in the review. \\
product-summary     &   Write a concise summary (maximum 5 sentences) that focuses on the product \\
                    &   categorization of the following Amazon review. Analyze the review to \\
                    &   determine the discussed product's categorization, referencing its features, \\
                    &   type, or purpose. Conclude with a categorization: \\
product-keyphrases  &   Generate keyphrases (max 5 phrases) that describe the product of the following \\
                    &   Amazon review. Focus on phrases that reflect the product's type or category. \\
topic-summary       &   Write a concise summary (maximum 5 sentences) that focuses on the topic or \\
                    &   theme of the following news article. Analyze the article to determine the main \\
                    &   topic or theme being discussed. Conclude with a topic categorization:  \\
topic-keyphrases    &   Generate keyphrases (max 5 phrases) that describe the topic of the following  \\
                    &   news article. Focus on phrases that reflect the main topic being discussed. \\ 
blurbs-summary      &   Schreibe eine prägnante Zusammenfassung (maximal 5 Sätze), die sich auf das \\  
                    &   Genre des folgenden Klappentextes konzentriert. Analysiere den Text, um das \\ 
                    &   Genre zu bestimmen. Schließe mit einer allgemeinen Genre-Kategorisierung ab. \\
blurbs-keyphrases   &   Generiere Schlüsselbegriffe (max 5 Begriffe), die das Genre des folgenden \\
                    &   Klappentextes beschreiben. Fokusiere dich auf Begriffe, die das allgemeine \\ 
                    &   Genre widerspiegeln. \\ 
stance-summary      &   Write a concise summary (maximum 5 sentences) that focuses on the stance \\
                    &   of the following Reddit post. Analyze the post to determine whether it is \\
                    &   Pro-Israel, Pro-Palestine, or Neutral. Conclude with a stance categorization:  \\
stance-keyphrases   &   Generate keyphrases (max 5 phrases) that describe the stance of the \\
                    &   following Reddit post. Focus on phrases that reflect the stance being discussed. \\
bias-summary        &   Write a concise summary (maximum 5 sentences) that focuses on the political \\
                    &   leaning of the following news article. Analyze the article to determine the main \\
                    &   political framing. Conclude with a political categorization:  \\
bias-keyphrases     &   Generate keyphrases (max 5 phrases) that describe the political leaning of \\
                    &   the following news article. Focus on phrases that reflect the political framing. \\ 
                    \bottomrule
\end{tabular}
\caption{Prompts used to rewrite the original documents. We use the same prompts for topics and frames.}
\label{tab:modification-prompts}
\end{table*}

\end{document}